\documentclass{article}
\usepackage{spconf,amsmath,graphicx,url}

\usepackage{multirow}

\usepackage{xcolor}

\newcommand{\Riva}[1]{\textcolor{black}{#1}}

\title{Is the U-Net Directional-Relationship Aware?}
\name{Mateus Riva$^{\star}$, Pietro Gori$^{\star}$, Florian Yger$^{\ddag}$, Isabelle Bloch$^{\dagger \star}$\thanks{Supported by Agence Nationale de la Recherche (ANR), project ANR-17-CE23-0021, São Paulo Research Foundation (FAPESP), projects 2017/50236-1 and 2015/22308-2, and ANR as part of the ``Investissements d'avenir" program, reference ANR-19-P3IA-0001 (PRAIRIE 3IA Institute).}}
\address{$^{\star}$ LTCI, Télécom Paris, Institut Polytechnique de Paris, France\\
    \{mateus.riva, pietro.gori\}@telecom-paris.fr\\
  $^\ddag$ LAMSADE, Université Paris-Dauphine, PSL Research University, France, 
    florian.yger@dauphine.fr\\
   $^{\dagger}$ Sorbonne Universit\'e, CNRS, LIP6, Paris, France,  isabelle.bloch@sorbonne-universite.fr }
\begin{document}
\ninept
\maketitle
\begin{abstract}
CNNs are often assumed to be capable of using contextual information about distinct objects (such as their directional relations) inside their receptive field. However, the nature and limits of this capacity has never been explored in full. We explore a specific type of relationship~-- directional~-- using a standard U-Net trained to optimize a cross-entropy loss function for segmentation. We train this network on a pretext segmentation task requiring directional relation reasoning for success and state that, with enough data and a sufficiently large receptive field, it succeeds to learn the proposed task. We further explore what the network has learned by analysing scenarios where the directional relationships are perturbed, and show that the network has learned to reason using these relationships.
\end{abstract}
\section{Introduction}
\label{sec:intro}

Convolutional neural networks (CNNs) and their variants are widely used with state-of-the-art results in many Computer Vision tasks. However, it is notably hard to ascribe reasoning properties to a CNN based solely on its performance, such as the capability for spatial reasoning in a structured scene. Despite the development of explainable artificial intelligence (XAI), most approaches trying to explain the predictions of CNN focus on local information only (regions or features involved in a decision)~\cite{Zhang2018-FITEE-visualinterpretabilitycnn} and not on the structure. However, reasoning capabilities would intuitively help CNNs avoid common pitfalls that hurt their generalization capability, such as some forms of dataset bias~\cite{Tommasi2017-BOOK-datasetbias} or their capacity of learning spurious correlations in the dataset while ignoring cues that are obvious to humans~\cite{Das2017-CVIU-networkattention}, such as structure in a scene.

Spatial relations have proved useful to assess the structure of a scene and to recognize the objects it contains~(see e.g.~\cite{Santoro2017-NIPS-relationalreasoning, Shaban2020-TMI-contextcnn},~\cite{IB:FSS-15} and the references therein) In this work, we focus on \textit{directional relationships}, where objects in a scene are distributed in specific directions and/or distances from others (e.g., ``the circle is 20 pixels to the left of the square, at the same height''). It is often assumed that CNNs have the inherent capacity for learning relevant relationships as long as they fit inside the receptive field~\cite{Shaban2020-TMI-contextcnn,Redmon2016-CVPR-YOLO, Yang2019-TGRS-roadcontextcnn,Mohseni2017-TMI-contextcnnbrain}. Other works assume that this capacity is not always guaranteed, and force or emphasise relationships using techniques external to the CNN~\cite{Santoro2017-NIPS-relationalreasoning, Zhou2018-ECCV-relationalreasoning}. Additionally, the use of certain performance measures do not put into evidence what was the reasoning process behind a decision. For all these reasons, it becomes hard to say if, when or how a given CNN learns a particular object relationship.

Differently from the aforementioned techniques, our work aims to explore the implicit assumption that a CNN can reason on relationships between objects in its receptive field, in a controlled manner. The objective of this paper is to determine if a basic U-Net, trained for a multi-object segmentation task with common loss functions, is capable of learning and using directional relationships between distinct objects to aid in their segmentation. To the best of the authors' knowledge, this scientific question has never been explored in-depth. We train the popular U-Net~\cite{ronneberger_u-net:_2015}, using commonly used hyperparameters, in a context where information on directional relations is key for perfect segmentation of objects of interest\Riva{; this experimental protocol, as well as the synthetic dataset used in its elaboration, are also both novel contributions}. Finally, we contribute to the growing field of neural network explainability by showcasing the performance of this network in such a context.  Our code is publicly available at \url{https://github.com/mateusriva/satann_synth}, \Riva{and supplementary experiments are available at \url{https://mateusriva.github.io}.}

\section{Related Work}
\label{sec:related_work}
Some recent works implicitly assume that CNNs inherently have relational reasoning capabilities. For instance, in their seminal paper YOLO, Redmon et al.~\cite{Redmon2016-CVPR-YOLO} mention that ``YOLO sees the entire image during training and test time so it implicitly encodes contextual information about classes''. Similar assertions are implicit in papers that link CNNs with larger receptive fields to usage of contextual information~\cite{Shaban2020-TMI-contextcnn, Yang2019-TGRS-roadcontextcnn, Mohseni2017-TMI-contextcnnbrain}. However, to the authors' knowledge, the extent of this implicit encoding has never been explored in full. We are particularly interested in the directional relationships, which provide semantics to the involved context (i.e. named relationships).

Recent relational reasoning works focus on explicit modeling. Some examples follow: Kamnitsas et al.~\cite{Kamnitsas2017-MIA-CRF} augment a 3D CNN with a Conditional Random Field to integrate local context during post-processing. Santoro et al.~\cite{Santoro2017-NIPS-relationalreasoning} and follow-up work by Zhou et al.~\cite{Zhou2018-ECCV-relationalreasoning} propose an extra MLP-based network module to improve CNN relational reasoning capabilities via self-attention. In a similar way, LSTM are widely used as an additional network in many works in image captioning, visual question answering. Janner et al.~\cite{Janner2018-TACL-spatialreasoning} mix text and visual information for solving relational reasoning based tasks, with the visual encoding being CNN-based, in a reinforcement learning scenario. Si et al.~\cite{Si2018-ECCV-skeletonactionspatial} perform skeleton-based action recognition with relational reasoning based on a graph neural network. Krishnaswamy et al.~\cite{Krishnaswamy2019-complexstructures} operate on the creation of a sequence of relational operations based on out-of-network search heuristics. However, these works fail to analyse the inherent capacity of CNNs for relational reasoning, by augmenting them with extra modules or replacing them entirely.

\section{Methods}
\label{sec:materials}
In this section, we present experimental methods for assessing the directional reasoning capabilities of the U-Net, by training on a pretext segmentation task that requires directional spatial reasoning for a correct answer. To this end, we present the synthetic Cloud of Structured Objects (CSO) dataset.

\subsection{The Cloud of Structured Objects Dataset}
\label{ssec:cso}
\begin{figure}[tb]
    \begin{minipage}[b]{.32\linewidth}
      \centering
      \centerline{\includegraphics[width=3.0cm]{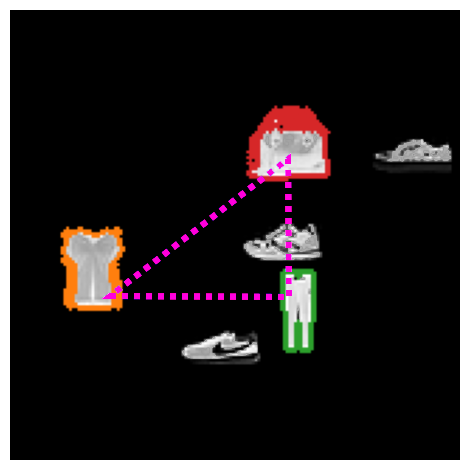}}
      \centerline{(a) T-Easy}\medskip
    \end{minipage}
    \hfill
    \begin{minipage}[b]{0.32\linewidth}
      \centering
      \centerline{\includegraphics[width=3.0cm]{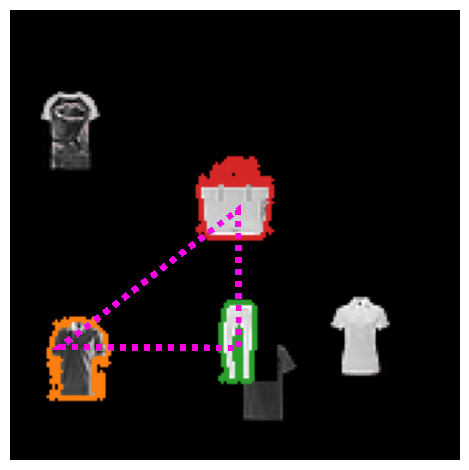}}
      \centerline{(b) T-Hard}\medskip
    \end{minipage}
    \hfill
    \begin{minipage}[b]{0.32\linewidth}
      \centering
      \centerline{\includegraphics[width=3.0cm]{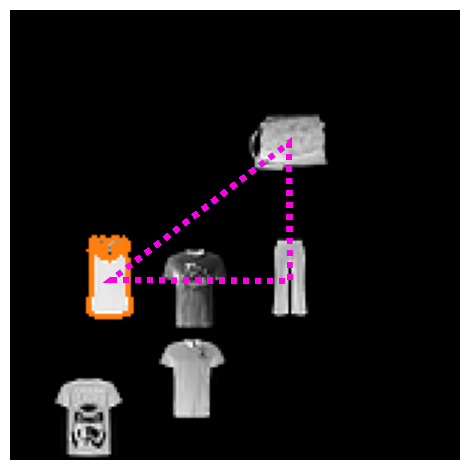}}
      \centerline{(c) T-Strict}\medskip
    \end{minipage}
    \caption{Examples of Cloud of Structured Objects images, for different configurations. The segmentation targets are highlighted. The directional relationships are represented by the dotted triangle.}
    \label{fig:cso_examples}
\end{figure}

The proposed Cloud of Structured Objects (CSO) dataset uses simple image datasets (such as the Fashion-MNIST~\cite{xiao_fashion_2017}) to generate a structured scene. A CSO data item is an image with objects of interest (OIs) of specific classes distributed in a structured way, along with several instances of a specified set of classes randomly distributed and called noise. The OIs (and only the OIs) are the segmentation targets, and are always at the foreground (i.e. they are never occluded by noise objects). OIs have a bounding box of size $28\times28$ pixels. We use a configuration (named ``T'') composed of three objects of interest, each belonging to a different class (specifically, ``shirts'', ``pants'', and ``bags'' from Fashion-MNIST). These objects form the vertices of a $48\times64\times80$ right-angled triangle, with its long leg laying horizontally, included in $160\times160$ 2D images (see Figure~\ref{fig:cso_examples}), thus determining the directional relationships between the objects. The entire OIs structure is translated by a random amount of pixels, drawn independently from a uniform distribution for each axis in the range of $[-32,32]$ pixels. We use the following noise distribution configurations:

\textbf{Easy}: three noise elements are added to the image, belonging to a different class from those of the objects of interest (``shoes'' in our experiments). Each individual OI is independently translated by a uniform random draw in the range of $[-16,16]$ pixels, resulting in a slightly imperfect triangle and adding noise to the directional relations.

\textbf{Hard}: similar to Easy, but the noise elements belong to the same class as one of the objects of interest (specifically, ``shirts''). Intuitively, the recognition and segmentation of the ``shirt'' OI must rely on its (imperfect) relationship with the other objects.



\textbf{Strict}: similar to Hard, but with no individual element positional noise (that is, the triangle is always perfect). Additionally, the noise elements are distributed only in the bottom-left region of the figure (inside a $80\times80$ square), and the triangle can be translated in the range of $[-40,40]$ pixels, so the absolute position information is useless in segmenting the OIs. The correct segmentation is only possible if the directional relationships between the objects are learned. Finally, only the class with noise (``shirts'') is considered as a segmentation target.
    
Examples of Fashion-MNIST-based CSO images of different configurations are displayed in Figure~\ref{fig:cso_examples}.

The harder CSO configurations present a joint segmentation and detection problem. Networks must learn to correctly detect and segment objects (a simple task), but must also learn to reason on which object is the correct one. A good segmentation of the correct object implies a high true-positive (TP) to false-negative (FN) rate. However, segmentation results that point to incorrect objects will result in a low true-positive (TP) to false-positive (FP) rate.

\subsection{U-Net Training}
\label{ssec:u-net_training}
The model training begins by choosing a CSO configuration and setting a size of the training and validation dataset $D$, from which $70\%$ is used for training, and the remaining $30\%$ for validation. 
We utilise a standard U-Net~\cite{ronneberger_u-net:_2015} with 4 levels. The receptive field at the bottleneck (respectively at the output) is $61\times61$ pixels (respectively $101 \times 101$ pixels)\footnote{Calculated using the \textit{receptivefield} library, available at \url{https://github.com/shelfwise/receptivefield}}, and thus can fit all OIs. We randomly initialise the models following He's initialisation~\cite{He2015-ICCV-init} with $5$ distinct seeds. The training/validation split is repeated randomly $5$ times.
For each CSO configuration, we train a network for $100$ epochs using an ADAM optimiser and cross-entropy loss function.

\begin{figure*}[htbp]
    \centering
    \begin{tabular}{c|ccc|c}
        Easy & \multicolumn{3}{c|}{Hard, Converging} & Hard, Non-Converging \\
        \includegraphics[width=0.17\linewidth]{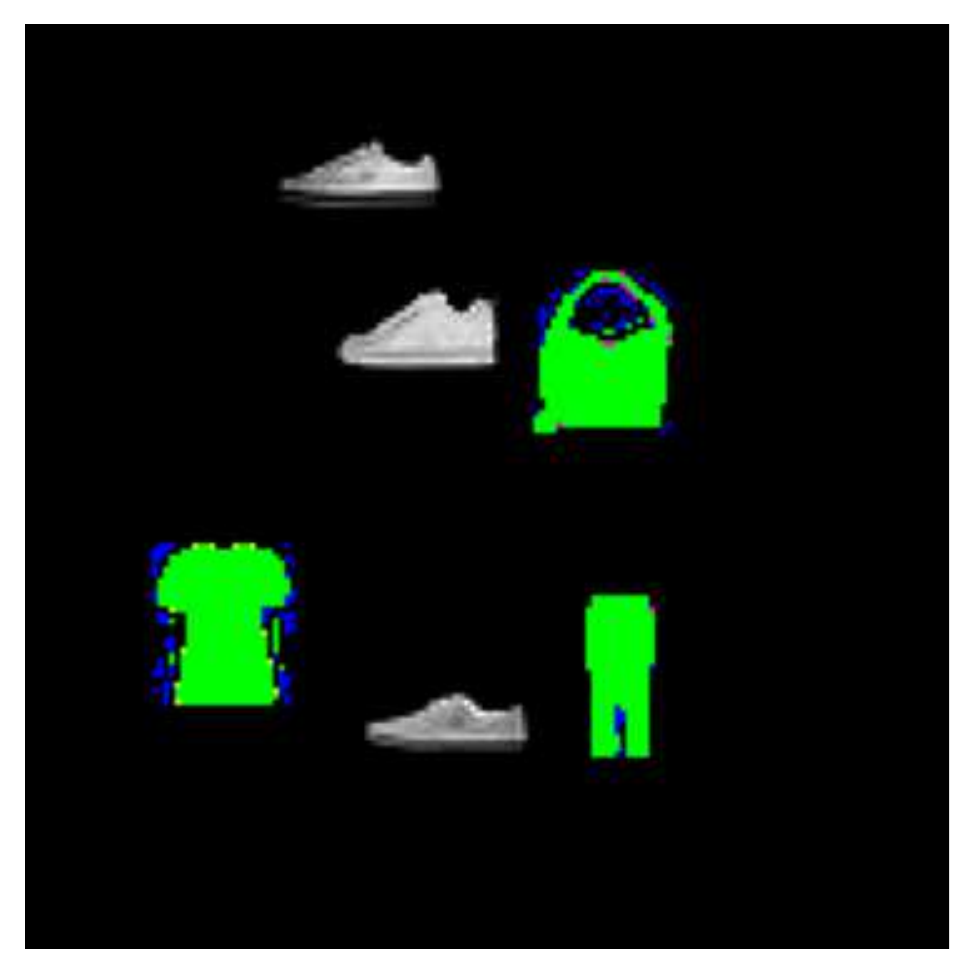} &
        \includegraphics[width=0.17\linewidth]{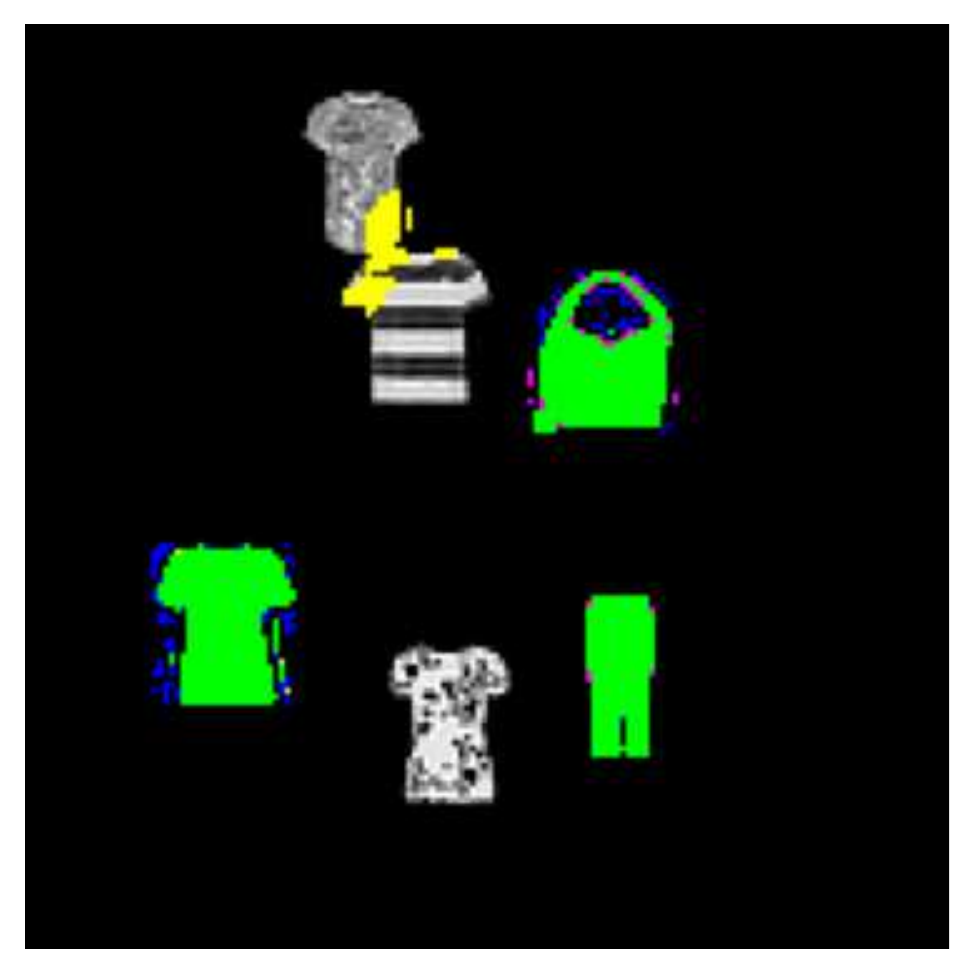} &
        \includegraphics[width=0.17\linewidth]{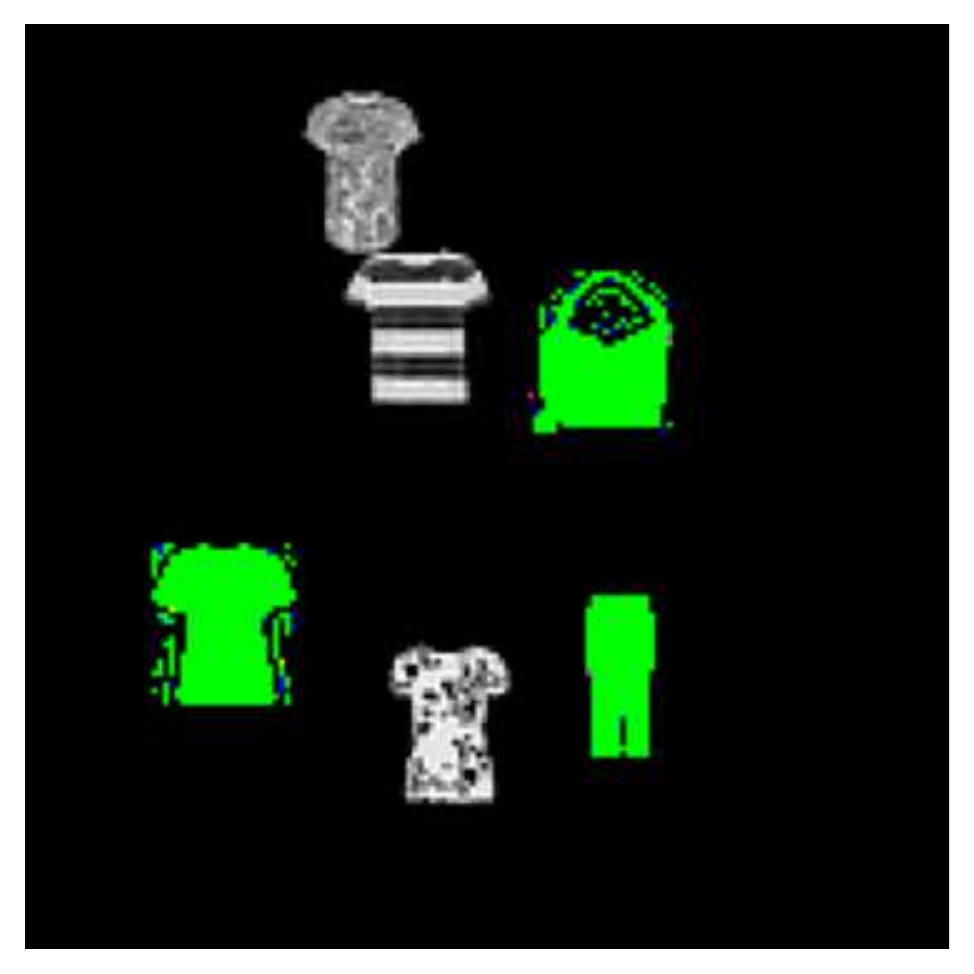} &
        \includegraphics[width=0.17\linewidth]{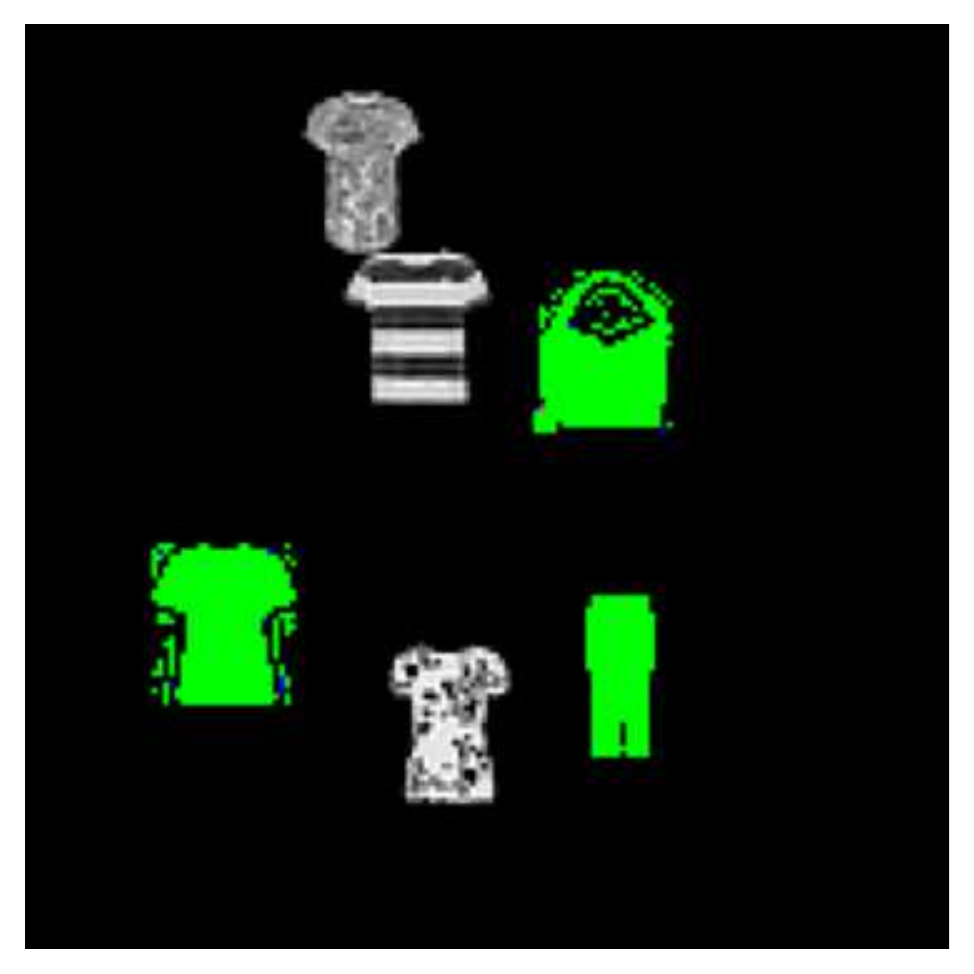} &
        \includegraphics[width=0.17\linewidth]{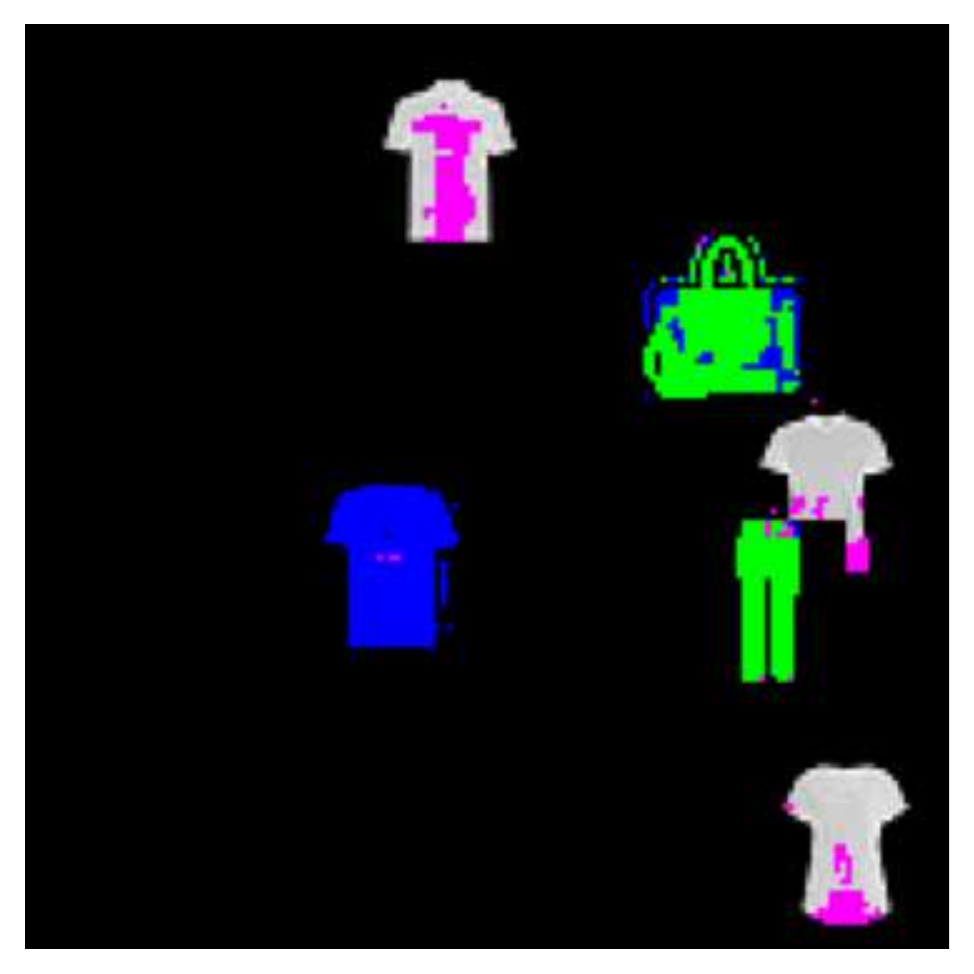} \\
        $D=100$ & $D=100$ & $D=1000$ & $D=1000$ & $D=100$ \\
        \includegraphics[width=0.17\linewidth]{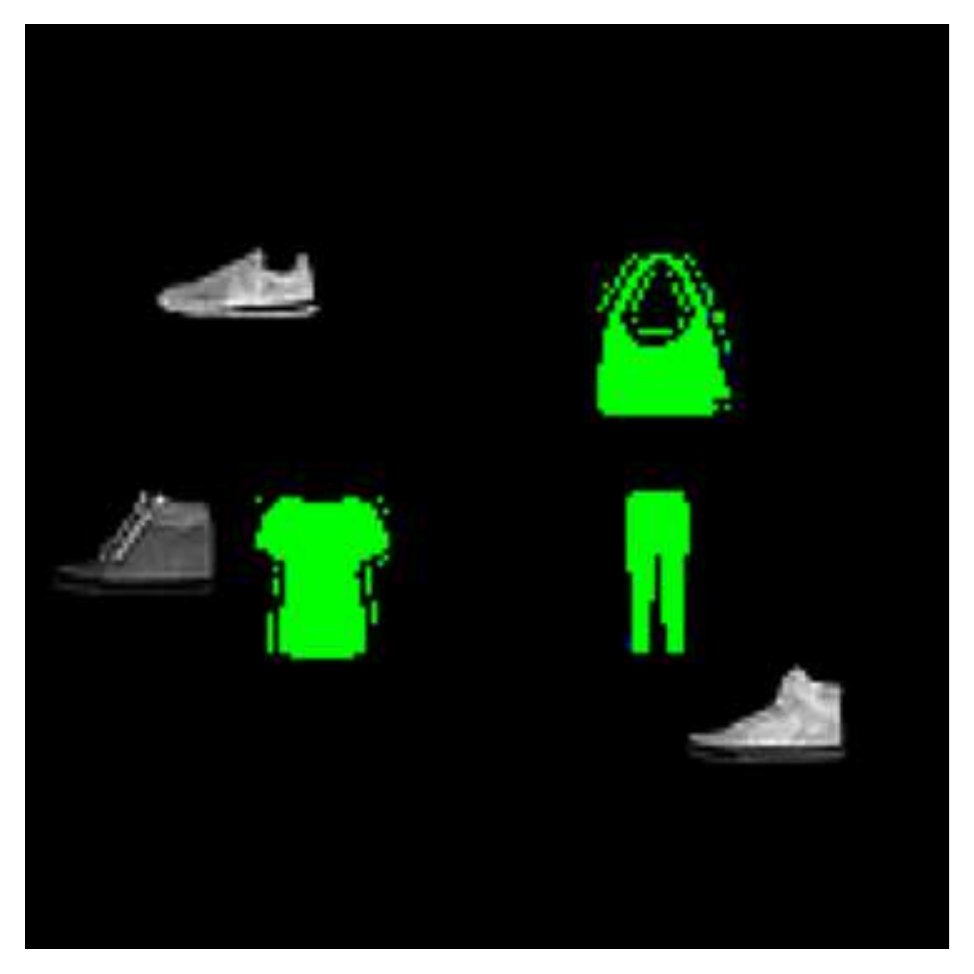} &
        \includegraphics[width=0.17\linewidth]{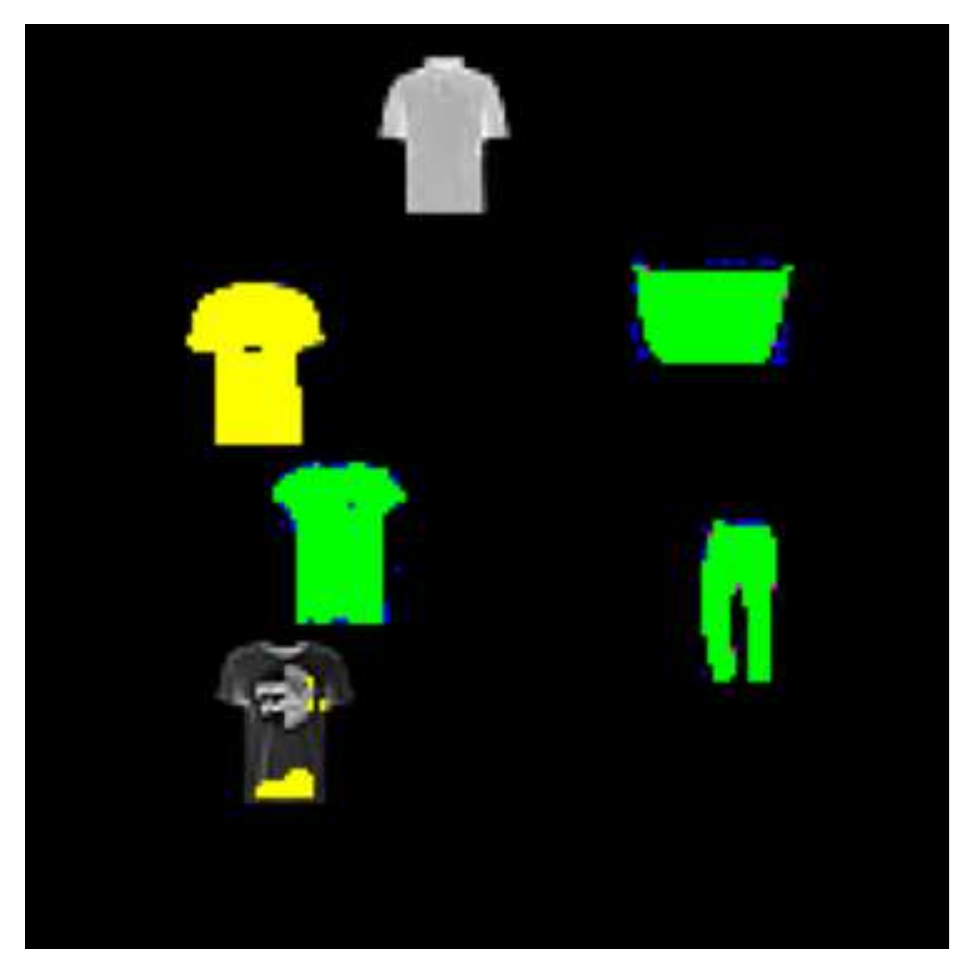} &
        \includegraphics[width=0.17\linewidth]{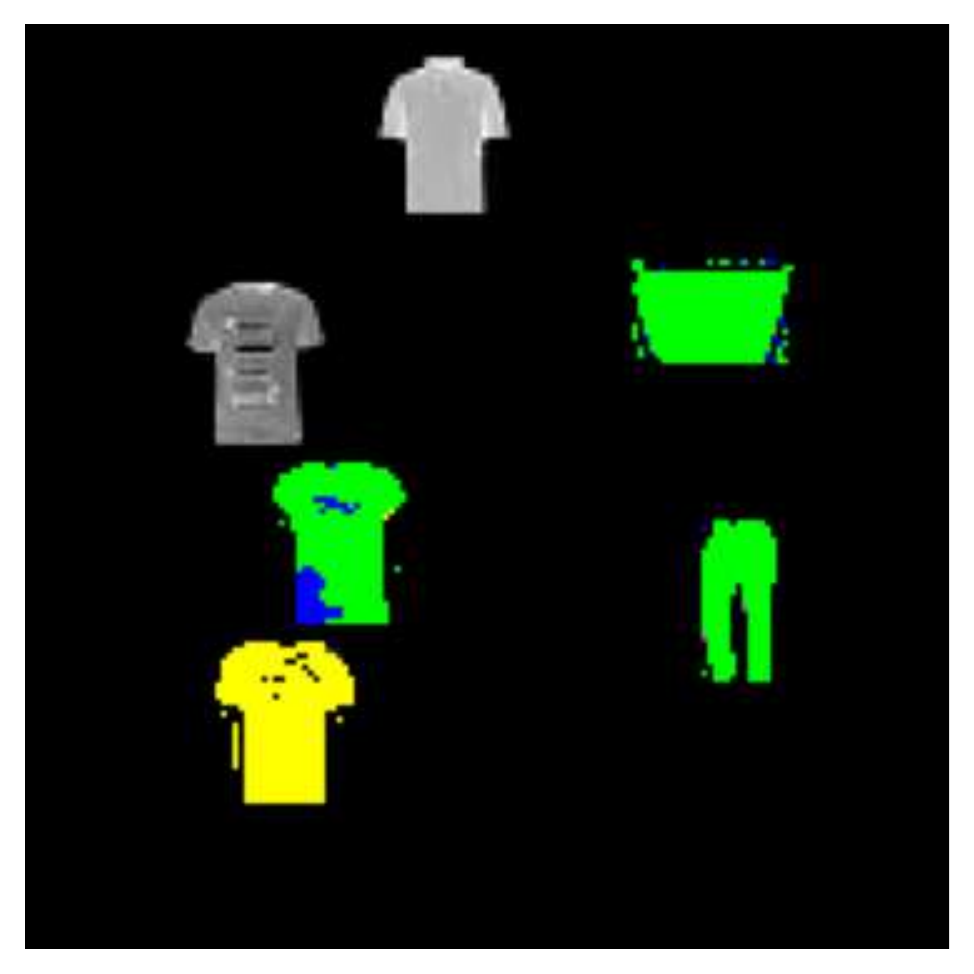} &
        \includegraphics[width=0.17\linewidth]{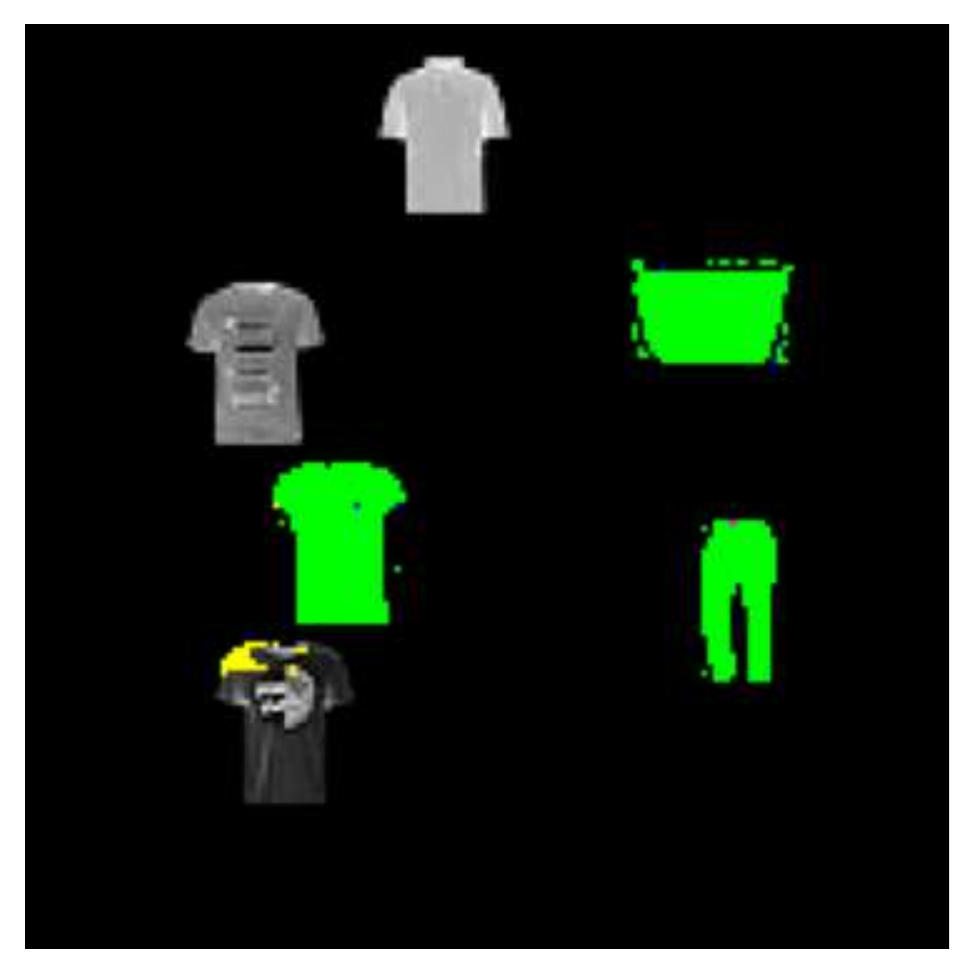} &
        \includegraphics[width=0.17\linewidth]{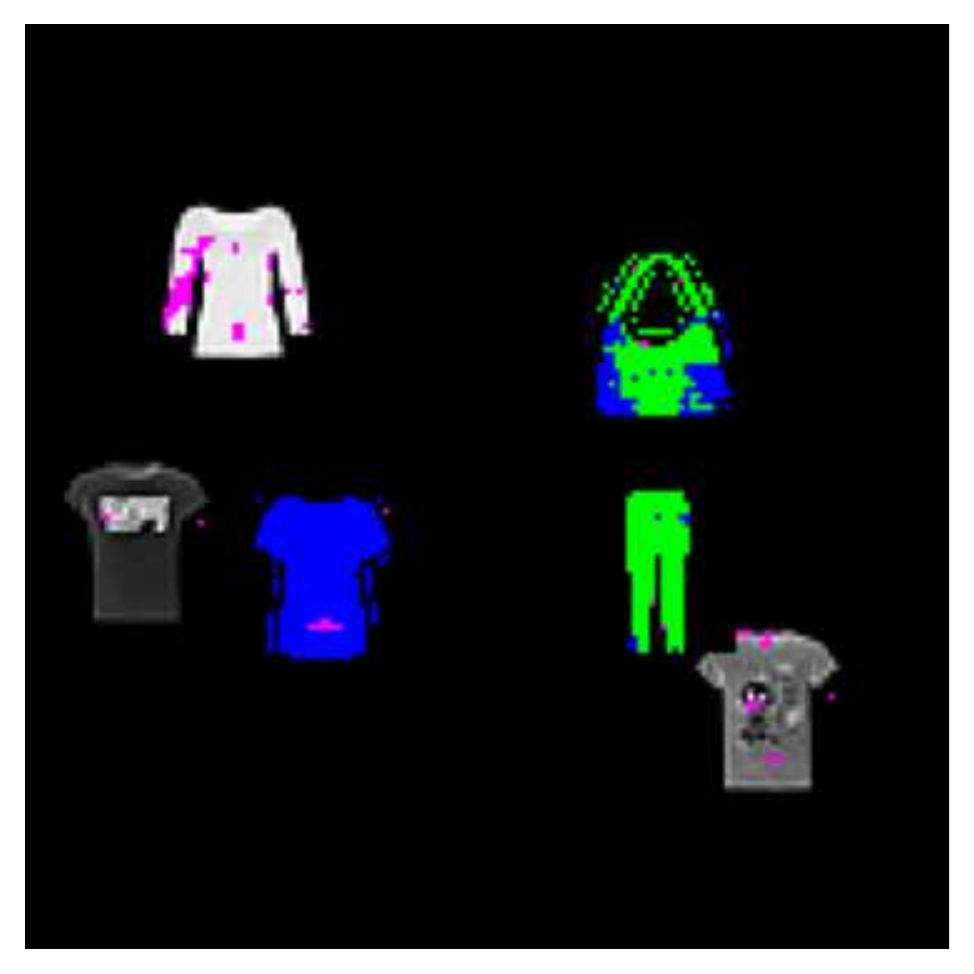} \\
        $D=1000$ & $D=100$ & $D=1000$ & $D=1000$ & $D=1000$ \\
    \end{tabular}
    \caption{Sample results of some of the trained models for the easier tasks. Green regions indicate true positives; blue regions indicate false negatives; yellow regions indicate false positives of the ``shirt'' class; magenta regions indicate false positives of the other classes. Noise around the OIs is inherited from the Fashion-MNIST dataset.}
    \label{fig:test_images_easy_hard}
\end{figure*}

\begin{figure*}[htbp]
    \centering
    \begin{tabular}{ccc|cc}
        \multicolumn{3}{c|}{Strict, Converging} & \multicolumn{2}{c}{Strict, Non-Converging} \\
        \includegraphics[width=0.17\linewidth]{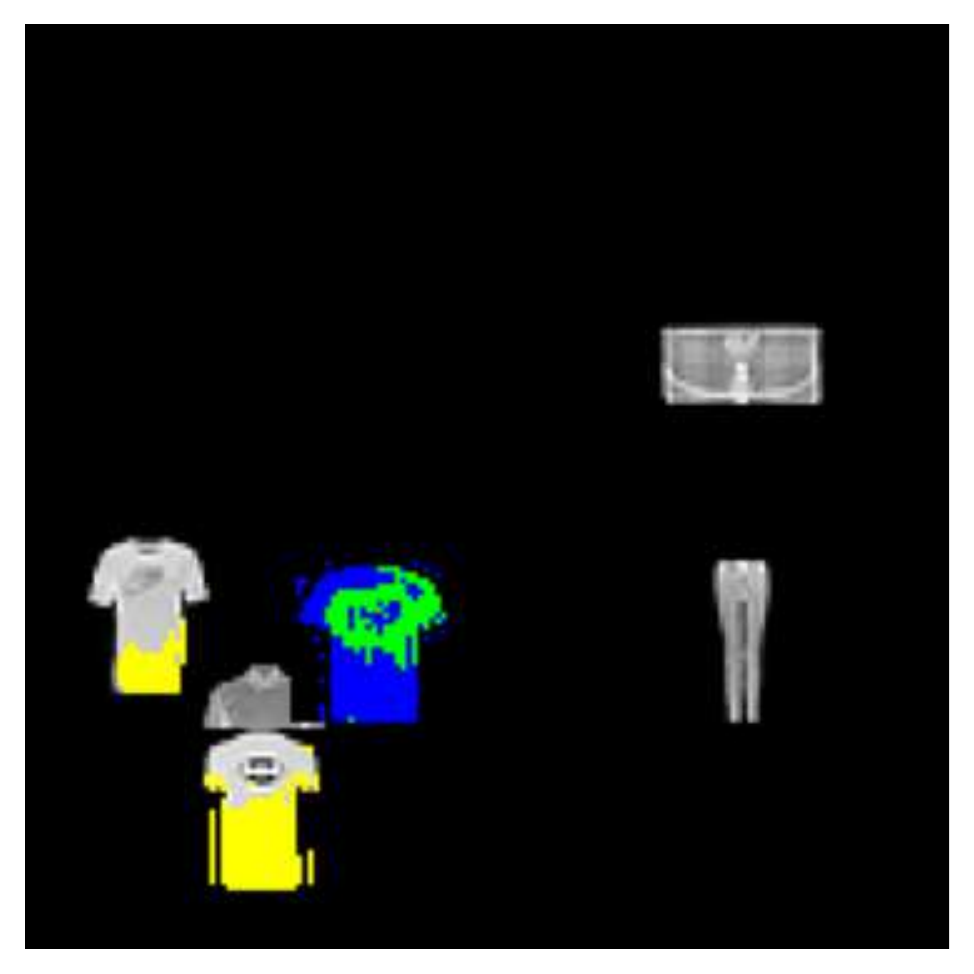} &
        \includegraphics[width=0.17\linewidth]{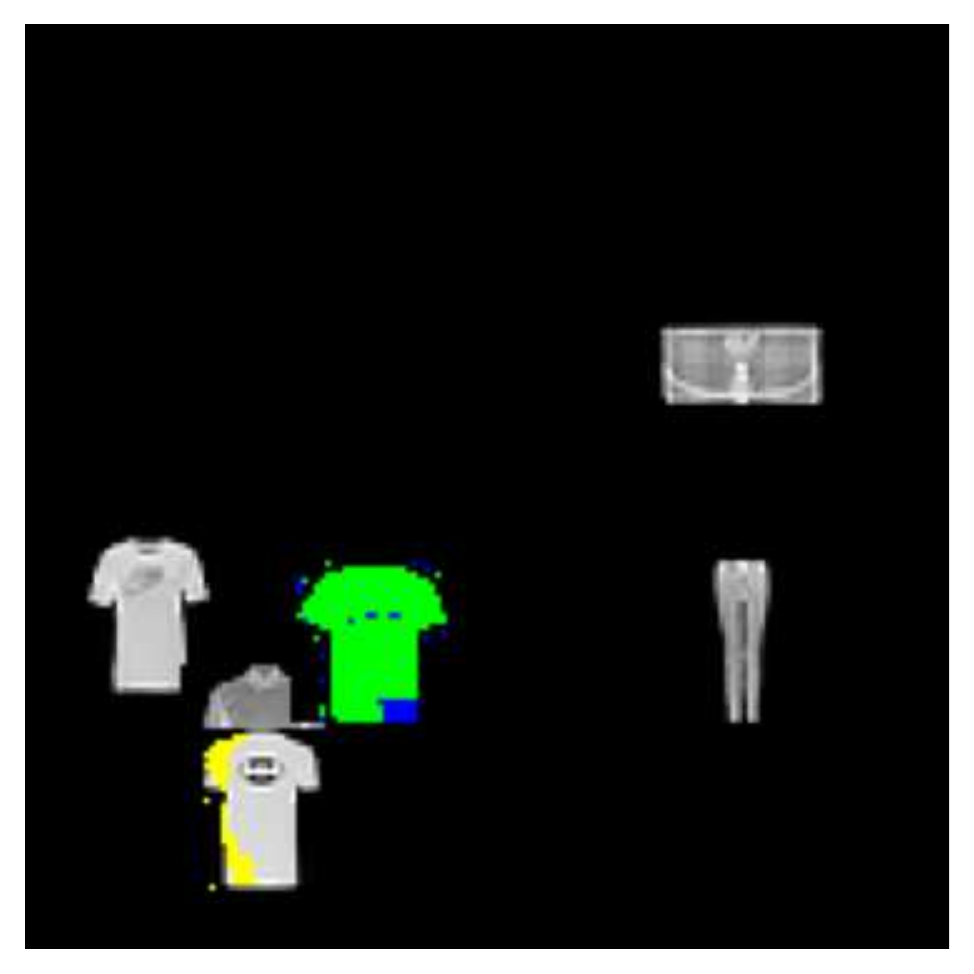} &
        \includegraphics[width=0.17\linewidth]{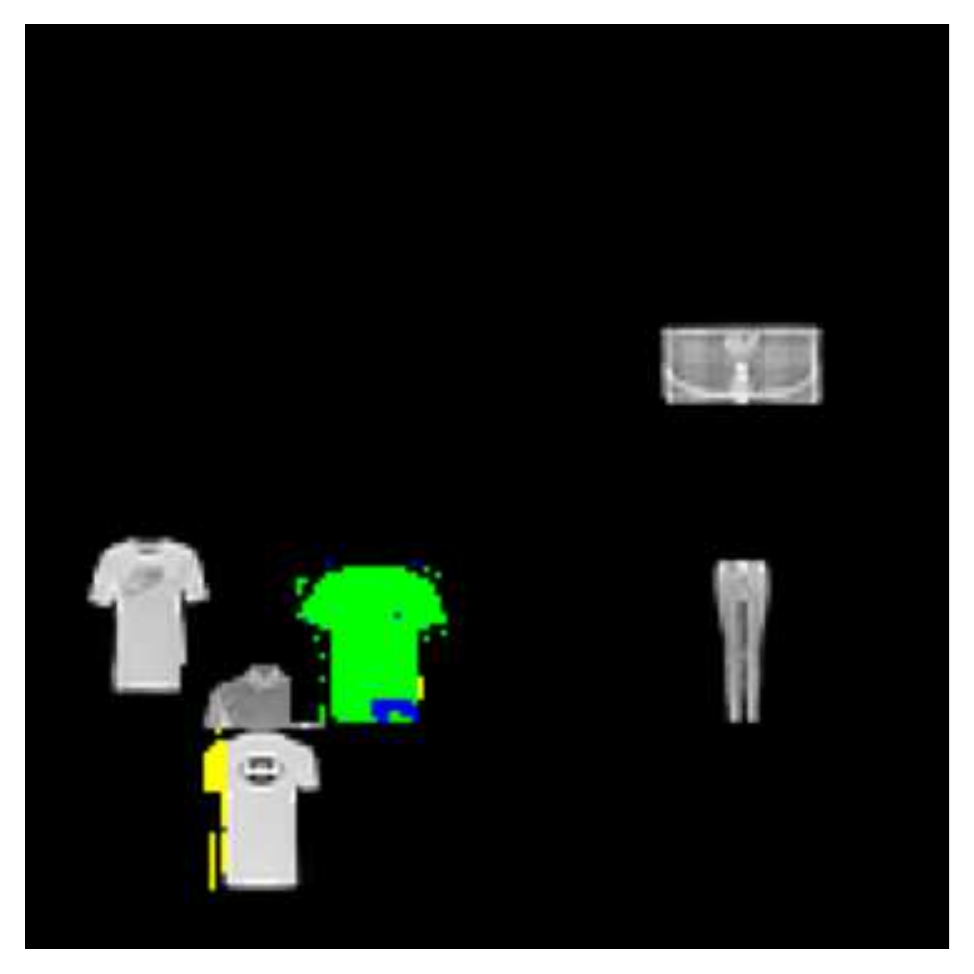} & 
        \includegraphics[width=0.17\linewidth]{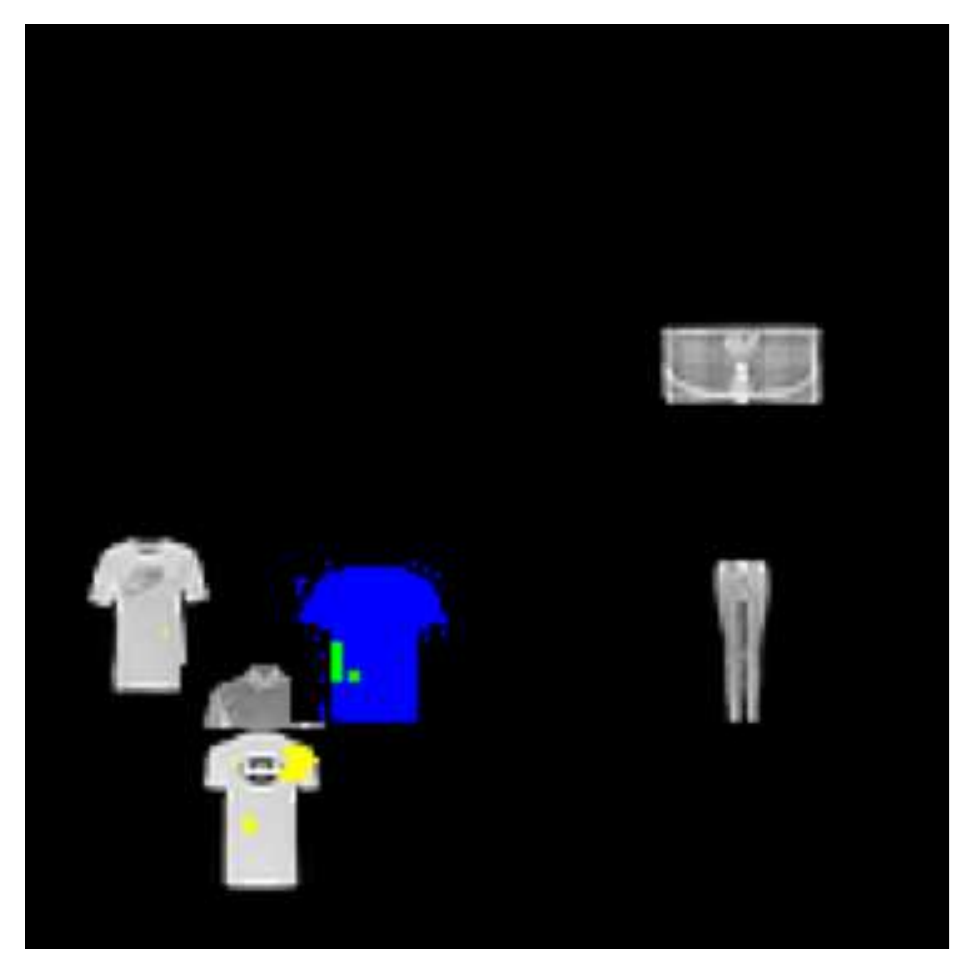} &
        \includegraphics[width=0.17\linewidth]{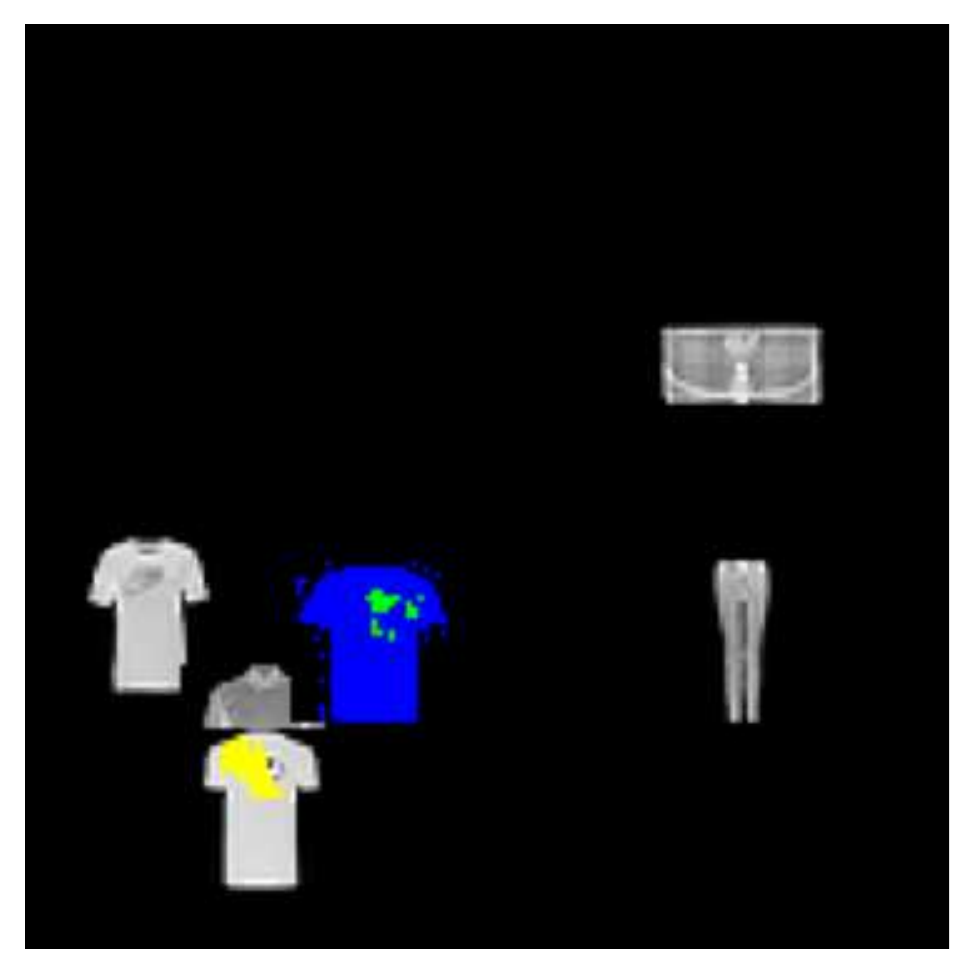} \\
        $D=1000$ & $D=5000$ & $D=10000$ & $D=1000$ & $D=5000$ \\
        \includegraphics[width=0.17\linewidth]{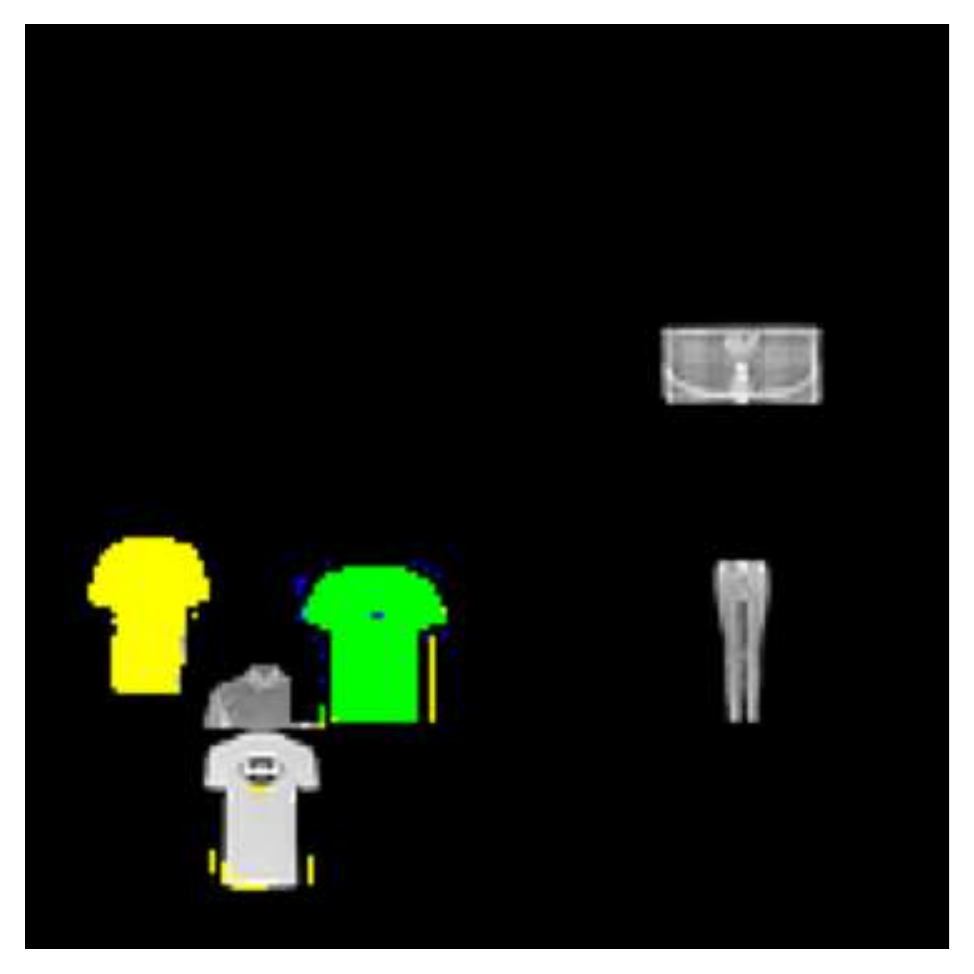} &
        \includegraphics[width=0.17\linewidth]{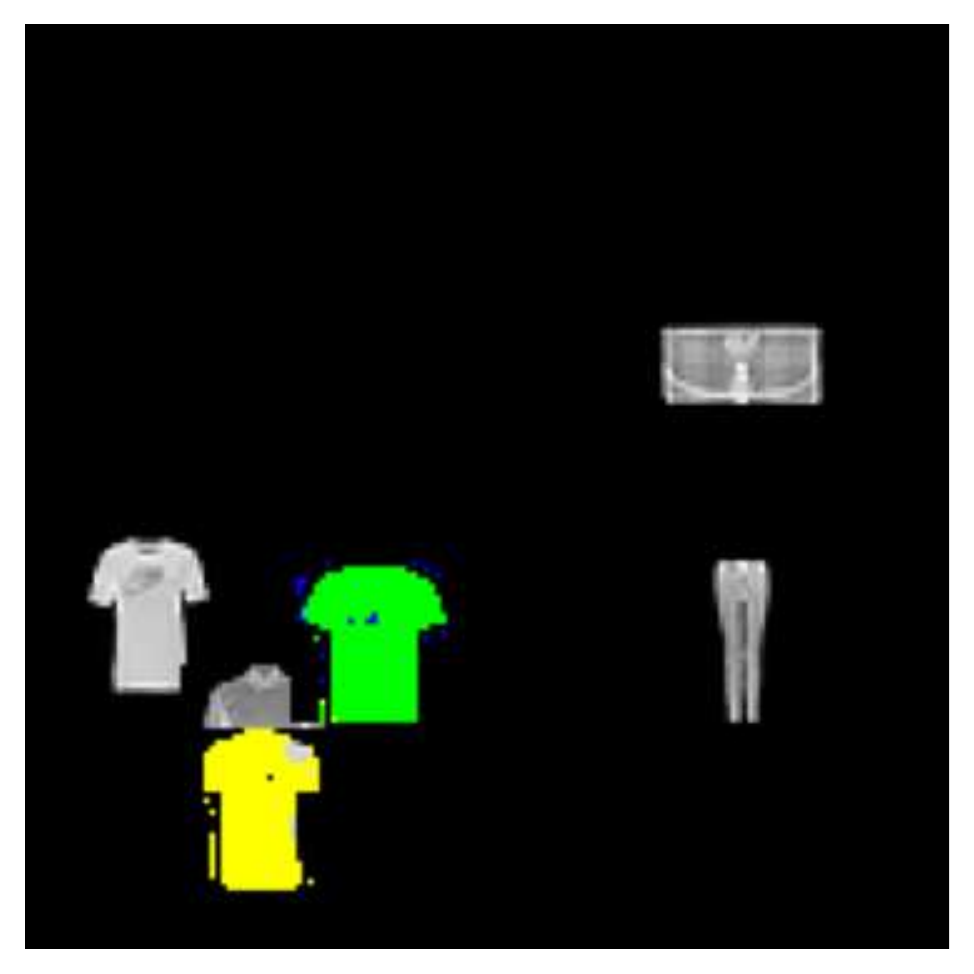} &
        \includegraphics[width=0.17\linewidth]{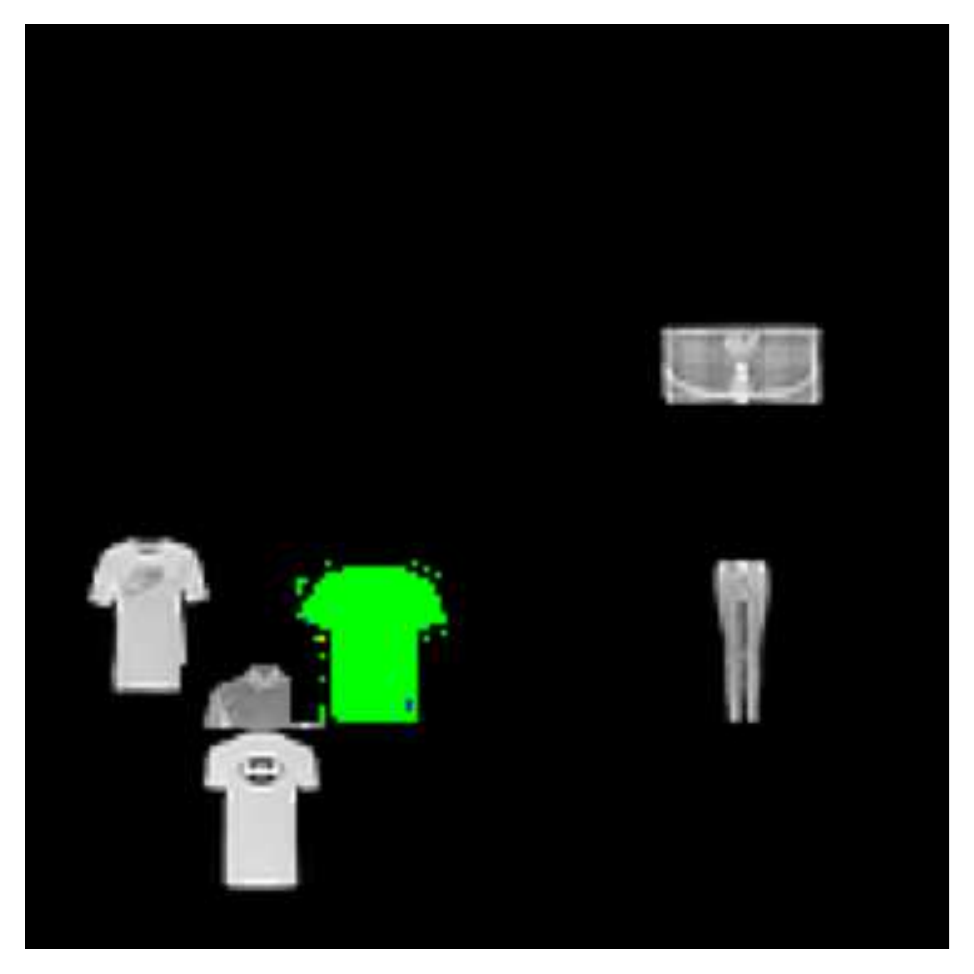} &
        \includegraphics[width=0.17\linewidth]{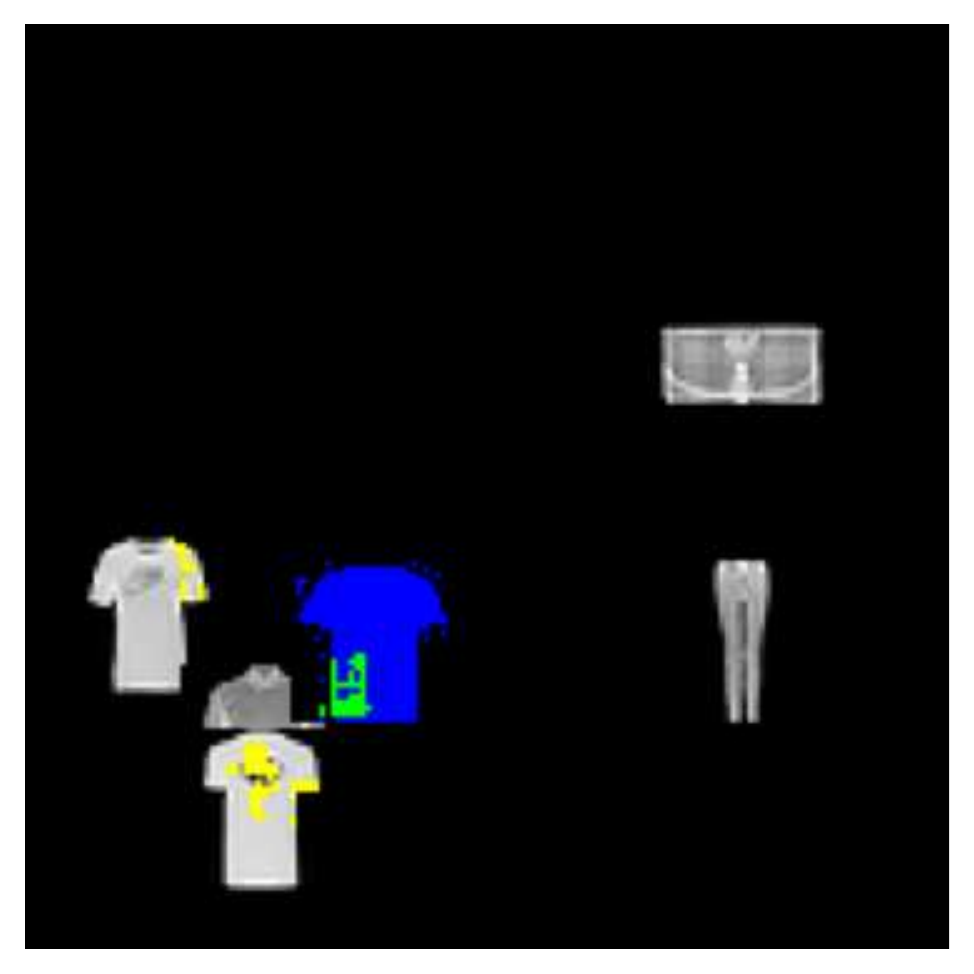} &
        \includegraphics[width=0.17\linewidth]{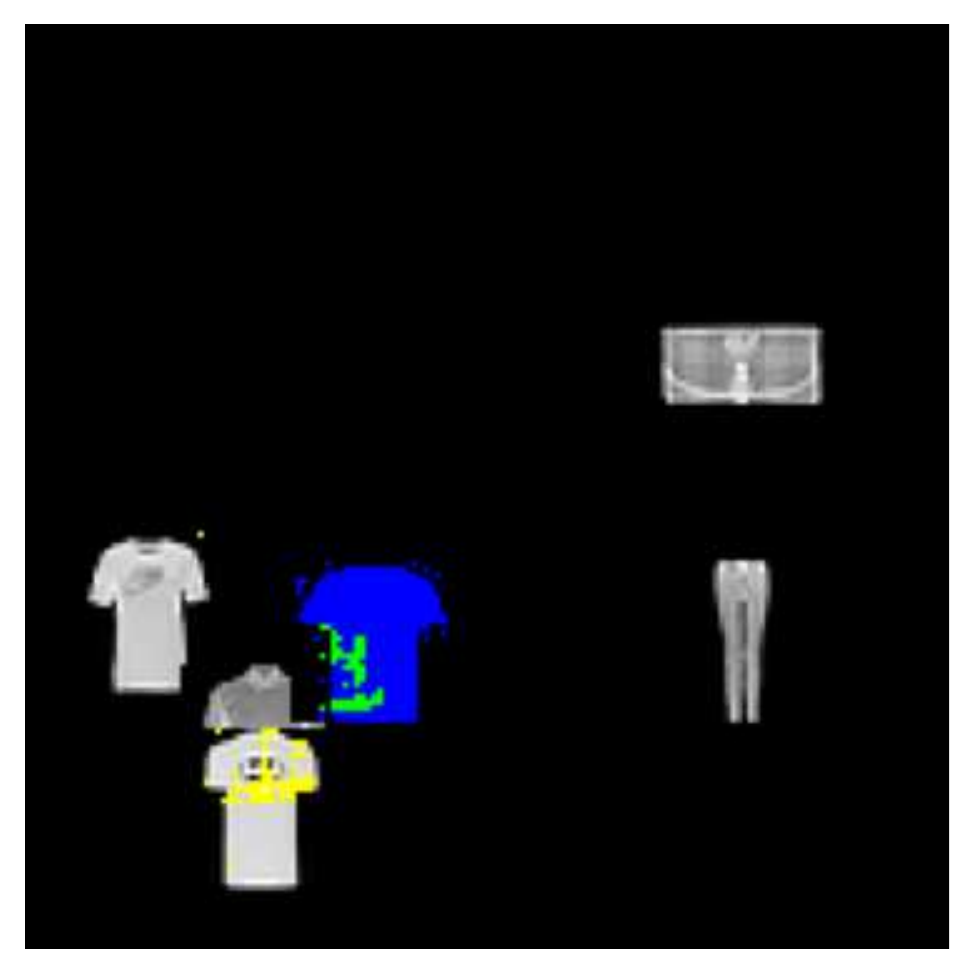} \\
        $D=1000$ & $D=5000$ & $D=50000$ & $D=10000$ & $D=50000$ \\
    \end{tabular}
    \caption{Sample results of some of the trained models for the strict task. The color code is the same as in Figure~\ref{fig:test_images_easy_hard}. 
    }
    \label{fig:test_images_strict}
\end{figure*}

To evaluate the models, we generate a test set containing $100$ new images of the same CSO configuration as the model, and use two measures: precision, defined as the per-pixel positive predictive value $\frac{TP}{TP+FP}$, and recall, defined as the per-pixel true positive rate $\frac{TP}{TP+FN}$. We compute the average test precision and recall for class ``shirt'', over all initializations where the model converged (defined as both precision and recall being above $0.5$). We also report how many of the trained models converged. Results are available in Table~\ref{tab:proof_of_learning}. Sample outputs are shown in Figures~\ref{fig:test_images_easy_hard} and~\ref{fig:test_images_strict}. 

\begin{table}[tb]
    \centering
    \begin{tabular}{c|c|cc|c}
        \multirow{2}{*}{Config.} & \multirow{2}{*}{$D$} & \multicolumn{2}{c|}{Class ``shirt''} & Conver-\\
        & & Precision & Recall & gences \\
        \hline
        \multirow{3}{*}{T-Easy} & 100 & $0.97 \pm 0.09$ & $0.95 \pm 0.11$ & 25/25\\
         & 1000 & $1.00 \pm 0.01$ & $1.00 \pm 0.03$ & 25/25\\
         & 10000 & $0.99 \pm 0.03$ & $0.99 \pm 0.04$ & 25/25\\
        \hline
        \multirow{3}{*}{T-Hard} & 100 & $0.83 \pm 0.23$ & $0.82 \pm 0.23$ & 24/25\\
         & 1000 & $0.95 \pm 0.14$ & $0.94 \pm 0.16$ & 24/25\\
         & 10000 & $0.98 \pm 0.11$ & $0.98 \pm 0.10$ & 25/25\\
        \hline
        \multirow{4}{*}{T-Strict} & 1000 & $0.65 \pm 0.32$ & $0.71 \pm 0.33$ & 6/25\\
         & 5000 & $0.79 \pm 0.29$ & $0.79 \pm 0.30$ & 14/25\\
         & 10000 & $0.87 \pm 0.19$ & $0.86 \pm 0.21$ & 21/25\\
         & 50000 & $0.91 \pm 0.14$ & $0.90 \pm 0.15$ & 22/25\\
    \end{tabular}
    \caption{Average precision and recall for the class ``shirt'', and standard deviation, for different dataset sizes and configurations, when the models converge.}
    \label{tab:proof_of_learning}
\end{table}

\subsection{Discussion}
We can see that properly segmenting and recognizing objects in the ``hard'' and ``strict'' cases is difficult with a small $D$. However, with enough data, the model learns to recognize the OIs. Lower precision scores in harder and/or small-dataset configurations point to the network being unable to completely avoid noise elements. The number of converging models shows that there is little guarantee of succeeding in the ``strict'' task without much more data than for the ``easy'' and ``hard'' tasks.

Analysing the example outputs of the networks, in Figures~\ref{fig:test_images_easy_hard} and~\ref{fig:test_images_strict}, sheds more light on the measures in Table~\ref{tab:proof_of_learning}. In Figure~\ref{fig:test_images_easy_hard}, for the ``Easy'' configuration, the network performs perfectly, which indicates that in a scenario without confusing noise objects (such as ``Hard'' and ``Strict''), the segmentation of the OIs is a simple task. For the converging models in the ``Hard'' configuration, most of the objects are correctly segmented, as can be seen by the high recall (and, correspondingly, true positives). However, scenarios where $D$ is smaller also run the risk of predicting noise objects. Finally, when the model fails to converge (rightmost column of Figure~\ref{fig:test_images_easy_hard}), we can see that it is still capable of predicting the ``bag'' and ``pants'' OIs, and simply omits all predictions of the class ``shirt''.

In Figure~\ref{fig:test_images_strict}, in the ``Strict'' scenario, non-converging models (on the two rightmost columns) still output some predictions, as the network had only a single segmentation target. However, they fail to properly detect and fully segment the correct shirt. In the converging cases (three leftmost columns), we can see the same expected tendency towards better segmentations when increasing data; it is clear that with enough data, the network can satisfy this task -- and thus, it must be capable of reasoning on directional relations.

\begin{figure*}[htpb]
    \centering
    \begin{tabular}{cccc|cccc}
        & \multicolumn{3}{c|}{``Hard'', $D=10000$} & \multicolumn{3}{c}{``Strict'', $D=50000$} & \\
        \rotatebox[origin=l]{90}{\textbf{Recall}} &
        \includegraphics[width=0.125\linewidth]{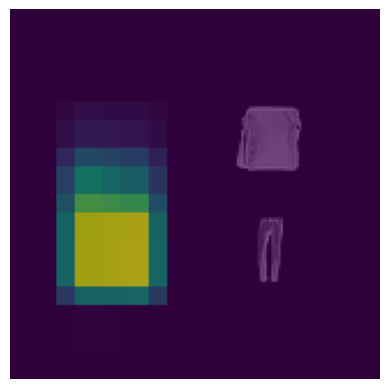} &
        \includegraphics[width=0.125\linewidth]{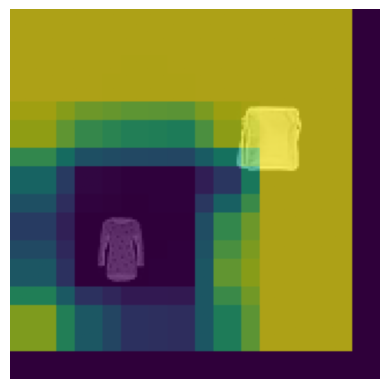} &
        \includegraphics[width=0.125\linewidth]{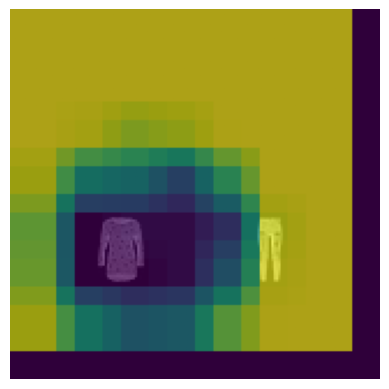} & 
        \includegraphics[width=0.125\linewidth]{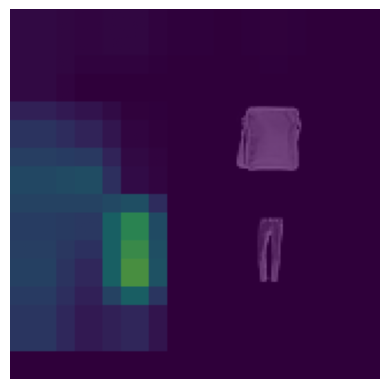} &
        \includegraphics[width=0.125\linewidth]{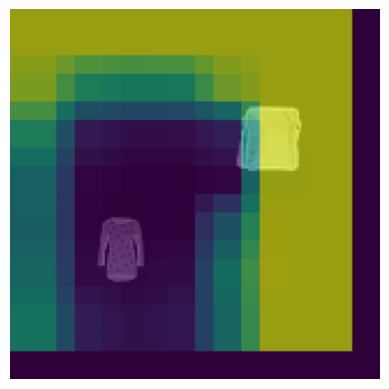} &
        \includegraphics[width=0.125\linewidth]{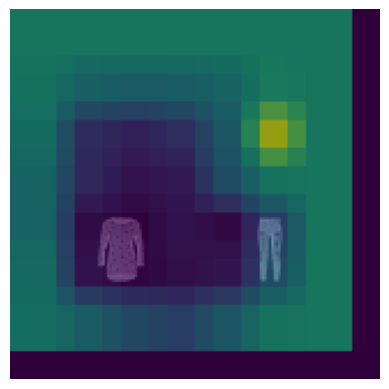} & 
        \multirow{2}{*}[1.4cm]{\includegraphics[scale=0.12]{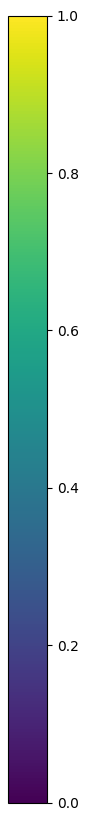}} \\
        \rotatebox[origin=l]{90}{\textbf{Precision}} &
        \includegraphics[width=0.125\linewidth]{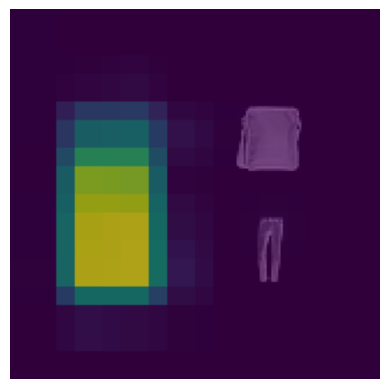} &
        \includegraphics[width=0.125\linewidth]{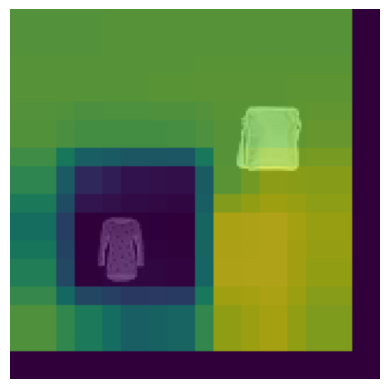} &
        \includegraphics[width=0.125\linewidth]{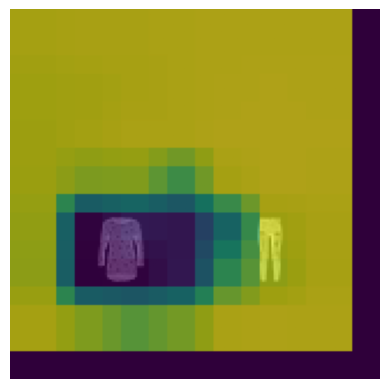} & 
        \includegraphics[width=0.125\linewidth]{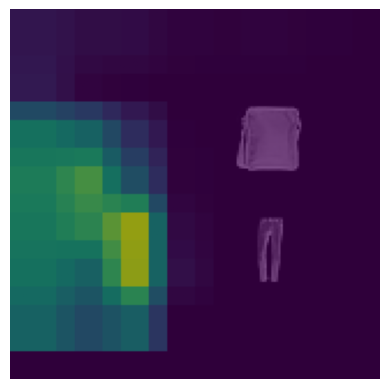} &
        \includegraphics[width=0.125\linewidth]{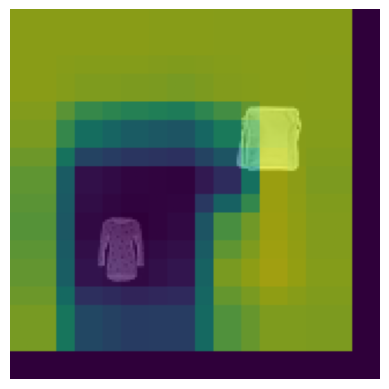} &
        \includegraphics[width=0.125\linewidth]{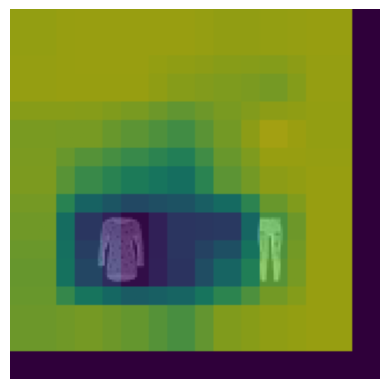} &  \\ 
        & Ref.: ``shirt'' & Ref.: ``pants'' & Ref.: ``bag'' & Ref.: ``shirt'' & Ref.: ``pants'' & Ref.: ``bag'' &
    \end{tabular}
    \caption{Precision and recall heatmaps of the class ``shirt'' when reference objects are slid across the image, for different configurations. Border effects are due to the sliding window limits.}
    \label{fig:heatmaps}
\end{figure*}

\section{Measuring Directional Relationship Awareness}
\label{sec:tests}

If the model learns to segment one OI by using another as a reference, we can expect that moving the reference around will affect the segmentation. To demonstrate this, we generate test images where each of the OIs, one at a time, is slid across the image using a stride of 20 pixels, while the other OIs remain fixed. The sliding OI is called the ``reference''. The ``reference'' is always at the foreground of the image. We compute the recall and precision of the segmentation of the ``shirt'' OI (even when it is used as ``reference''). For all positions of the reference, $20$ images are generated with the triangle perfectly centered and noise distributed according to the considered configuration.

We then build a heatmap, where its value at a specific location $(x,y)$ is the averaged evaluation measure (either precision or recall) of the class ``shirt'' when the reference object is at position $(x,y)$. In the ``hard'' and ``strict'' configurations, if the network has learned to use other classes for the segmentation of the OI, we expect to see poor performance when the references are not positioned at their expected places. 

Figure \ref{fig:heatmaps} show the resulting heatmaps on the two largest datasets for the ``Hard'' and ``Strict'' configurations. To facilitate interpretation, the heatmaps are overlayed on a dummy image showing the centered OI structure, and the reference is not displayed.


In the first and fourth columns, where the ``shirt'' itself is slid across the image, we can see that its segmentation can only happen in a specific region of the image. This may be due to the network needing the other OIs to segment the ``shirt'', learning the absolute positions where the ``shirt'' can be found, or a combination of both. In the second and fifth columns, we can see that the position of the ``pants'' does not affect the recall of the ``shirt'' (except when the ``pants'' occlude the ``shirt''); the precision of the ``shirt'', however, benefits from the proper positioning of the ``pants'' (highest values of precision in the heatmap), implying that it plays some role in allowing the network to avoid segmenting the wrong ``shirts''. Finally, in the third and sixth columns, we see that the same observations made for the ``pants'' as the reference are true for the ``bag'', with the notable exception of the recall in the ``strict'' case (sixth column, top image). In that case, the recall is remarkably diminished when the bag is not perfectly placed. All of this is a further evidence that the U-Net has learned to use other objects when reasoning about the segmentation of the shirt OI.

\section{CONCLUSIONS}
From the experiments shown, it can be reasonably concluded that the U-Net is indeed capable of reasoning between different objects in its receptive field, and using directional relationships to ensure proper segmentation. When trained on a task requiring directional relational reasoning, a simple U-Net trained with a cross-entropy loss function was capable of attaining satisfactory results, when enough data were supplied. {Our tests also show that} disturbing the directional relationships in test data directly results in underperformance, helping to explain the nature of the relationships learned by the network.

This work is but a first step towards improving CNN explainability by better understanding how basic CNNs can reason about relationships between objects contained in their receptive fields. We have demonstrated that a CNN \textbf{can} learn to contextualise objects~-- specifically, it can learn directional spatial relationships --~in its receptive field, alongside putting into evidence the data hunger inherent to complicated reasoning tasks. Further works will aim at exploring this question in different directions:
(i) what are the details of the relationship learning process?
(ii) can relationship learning be accelerated?
(iii) will accelerating relationship learning result in better-performing networks or lessen training data hunger?
(iv) what are the limits of relational reasoning (such as behavior when facing overly narrow or sparse receptive fields)?


\bibliographystyle{IEEEbib}
\bibliography{refs}

\end{document}